\documentclass{article}

\usepackage{algorithm}
\usepackage{algorithmic}
\usepackage{amsmath, amssymb}
\usepackage{bm}
\usepackage{booktabs}
\usepackage{graphicx}
\usepackage{hyperref}
\usepackage{mathtools}
\usepackage{multirow}
\usepackage{natbib}
\usepackage[separate-uncertainty=true,multi-part-units=repeat]{siunitx}
\usepackage{xcolor}
\usepackage{xspace}

\usepackage{adjustbox}
\usepackage{hyperref}
\usepackage{url}

\hypersetup{
  pdfinfo={
    Title={Mode Normalization},
    Author={Lucas Deecke, Iain Murray, Hakan Bilen}
  }
}

\newcommand{\mn}{MN\xspace}
\newcommand{\mgn}{MGN\xspace}

\newcommand{\E}{\small$\pm$\,}
\newcommand{\ul}{\underline}

\definecolor{velvet}{HTML}{5544BA}

\definecolor{brown}{HTML}{8A4319}

\makeatletter
\DeclareRobustCommand\onedot{\futurelet\@let@token\@onedot}
\def\@onedot{\ifx\@let@token.\else.\null\fi\xspace}
\def\eg{\emph{e.g}\onedot} 
\def\ie{\emph{i.e}\onedot} 
\def\cf{\emph{c.f}\onedot}

\makeatother

\title{Mode Normalization}

\author{Lucas Deecke, Iain Murray, Hakan Bilen \\
University of Edinburgh \\
\texttt{\{l.deecke,i.murray,hbilen\}@ed.ac.uk}}

\begin{document}

\maketitle

\begin{abstract}
Normalization methods are a central building block in the deep learning toolbox. They accelerate and stabilize training, while decreasing the dependence on manually tuned learning rate schedules. When learning from multi-modal distributions, the effectiveness of batch nor\-mal\-iza\-tion (BN), arguably the most prominent normalization method, is reduced. As a remedy, we propose a more flexible approach:\ by extending the normalization to more than a single mean and variance, we detect modes of data on-the-fly, jointly normalizing samples that share common features. We demonstrate that our method outperforms BN and other widely used normalization techniques in several experiments, including single and multi-task datasets.
\end{abstract}

\section{Introduction}
\label{sec:introduction}

A fundamental challenge in optimizing deep learning models is the continuous change in input distributions at each layer, complicating the training process. Normalization methods, such as batch normalization (BN) \citep{ioffe15} are aimed at overcoming this issue\,---\,often referred to as internal covariate \mbox{shift \citep{shimodaira2000improving}}.\footnote{Note that the underlying mechanisms are still being explored from a theoretical perspective, see \citet{kohler18,santurkar18}.} When applied successfully in practice, BN enables the training of very deep networks, shortens training times by supporting larger learning rates, and reduces sensitivity to parameter initializations. As a result, BN has become an integral element of many state-of-the-art machine learning techniques~\citep{he16,silver2017mastering}.

Despite its great success, BN has drawbacks due to its strong reliance on the mini-batch statistics. While the stochastic uncertainty of the batch statistics acts as a regularizer that can boost the robustness and generalization of the network, it also has significant disadvantages when the estimates of the mean and variance become less accurate. In particular, heterogeneous data \citep{bilen2017universal} and small batch sizes \citep{ioffe17,wu18} are reported to cause inaccurate estimations and thus have a detrimental effect on models that incorporate BN\@. For the former, \citet{bilen2017universal} showed that when training a deep neural network on images that come from a diverse set of visual domains, each with significantly different statistics, then BN is not effective at normalizing the activations with a single mean and variance. 

\begin{figure}[t]
\begin{center}
\centerline{\includegraphics[width=\textwidth]{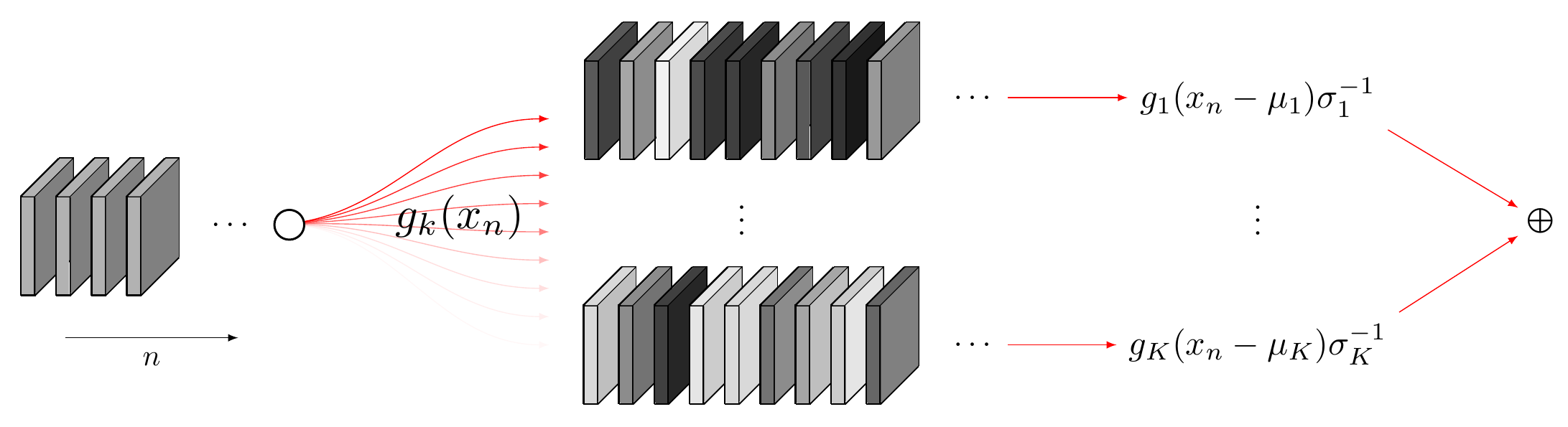}}
\end{center}
\caption{In mode normalization, incoming samples $\{x_n\}_{n=1,\dots,N}$ are weighted by a set of gating functions $\{g_k\}_{k=1, \dots,K}$. Gated samples contribute to component-wise estimators $\mu_k$ and $\sigma_k$, under which the data is normalized. After a weighted summation, the batch is passed on to the next layer. Note that during inference, estimators are computed from running averages instead.} \label{fig:sketch}
\end{figure}

In this paper we relax the assumption that the entire mini-batch should be normalized with the same mean and variance. We propose a novel normalization method, mode normalization (\mn), that first assigns samples in a mini-batch to different modes via a gating network, and then normalizes each sample with estimators for its corresponding mode (see Figure~\ref{fig:sketch}). We further show that \mn can be incorporated into other normalization techniques such as group normalization (GN) \citep{wu18} by learning which filters should be grouped together. The proposed methods can easily be implemented as layers in standard deep learning libraries, and their parameters are learned jointly with the other parameters of the network in an end-to-end manner. We evaluate \mn on multiple classification tasks and demonstrate that it achieves a consistent improvement over BN and GN.

In Section~\ref{sec:background}, we present how this paper is related to previous work. We then review BN and GN, and introduce our method in Section~\ref{sec:method}. The proposed methods are evaluated on multiple benchmarks in Section~\ref{sec:experiments}, and our findings are summarized in Section~\ref{sec:conclusion}.

\section{Related work}
\label{sec:background}

\paragraph{Normalization.} Normalizing input data~\citep{lecun1998efficient} or initial weights of neural networks~\citep{glorot2010understanding} are known techniques to support faster model convergence, and were studied extensively in previous work. More recently, normalization has been evolved into functional layers to adjust the internal activations of neural networks. Local response normalization (LRN)~\citep{lyu2008nonlinear,jarrett2009best} is used in various models~\citep{krizhevsky12,sermanet2014overfeat} to perform normalization in a local neighborhood, and thereby enforce competition between adjacent pixels in a feature map. BN~\citep{ioffe15} implements a more global normalization along the batch dimension. In contrast to LRN, BN requires two distinct train and inference modes. At training time, samples in each batch are normalized with the batch statistics, while during inference samples are normalized using precomputed statistics from the training set. Small batch sizes or heterogeneity can lead to inconsistencies between training and test data. Our proposed method alleviates such issues by better dealing with different modes in the data, simultaneously discovering these and normalizing the data accordingly.

Several recent normalization methods~\citep{ba16,ulyanov17,ioffe17} have emerged that perform normalization along the channel dimension~\citep{ba16}, or over a single sample~\citep{ulyanov17} to overcome the limitations of BN\@. \citet{ioffe17} proposes a batch renormalization strategy that clips gradients for estimators by using a predefined range to prevent degenerate cases. While these methods are effective for training sequential and generative models respectively, they have not been able to reach the same level of performance as BN in supervised classification. Simultaneously to these developments, BN has started to attract attention from theoretical viewpoints \citep{kohler18,santurkar18}.

More recently, Wu and He~\citep{wu18} have proposed a simple yet effective alternative to BN by first dividing the channels into groups and then performing normalization within each group. The authors show that group normalization (GN) can be coupled with small batch sizes without any significant performance loss, and delivers comparable results to BN when the batch size is large. We build on this method in Section~\ref{sec:methodmn}, and show that it is possible to automatically infer filter groupings. An alternative normalization strategy is to design a data independent reparametrization of the weights in a neural network by implicitly whitening the representation obtained at each layer \citep{desjardins2015natural,arpit2016normalization}. While these methods show promising results, they do not generalize to arbitrary non-linearities and layers.

\paragraph{Mixtures of experts.} Mixtures of experts (MoE)~\citep{jacobs91,jordan94} are a family of models that involve combining a collection of simple learners to split up the learning problem. Samples are thereby allocated to differing subregions of the model that are best suited to deal with a given example. There is a vast body of literature describing how to incorporate MoE with different types of expert architectures such as SVMs~\citep{collobert02}, Gaussian processes~\citep{tresp01}, or deep neural networks~\citep{eigen2013learning,shazeer17}. Most similar to ours, \citet{eigen2013learning} propose to use a different gating network at each layer in a multilayer network to enable an exponential number of combinations of expert opinions. While our method also uses a gating function at every layer to assign the samples in a mini-batch to separate modes, it differs from the above MoE approaches in two key aspects: (i.)~we use the assignments from the gating functions to normalize the data within a corresponding mode, (ii.)~the normalized data is forwarded to a common module (\ie a convolutional layer) rather than to multiple separate experts.

Our method is also loosely related to Squeeze-and-Excitation Networks~\citep{hu2018squeeze}, that adaptively recalibrate channel-wise feature responses with a gating function. Different to their approach, we use the outputs of the gating function to normalize the responses within each mode.

\paragraph{Multi-domain learning.} Our approach also relates to methods that pa\-ra\-met\-rize neural networks with domain-agnostic and specific layers, and transfer the agnostic parameters to the analysis of very different types of images \citep{bilen2017universal,rebuffi17,rebuffi18}. In contrast to these methods, which require the supervision of domain knowledge to train domain-agnostic parameters, our method can automatically learn to discover modes both in single and multi-domain settings, without any supervision.

\section{Method}\label{sec:method}

We first review the formulations of BN and GN in Section~\ref{sec:methodbn}, and introduce our method in Section~\ref{sec:methodmn}.

\subsection{Batch and group normalization}\label{sec:methodbn}
Our goal is to learn a prediction rule $f \colon \mathcal X \to \mathcal Y$  that infers a class label $y \in \mathcal Y$ for a previously unseen sample $x \in \mathcal X$. For this purpose, we optimize the parameters of $f$ on a training set $\{x_i\}_{i=1,\dots,N_d}$ for which the corresponding label information $\{y_i\}_{i=1,\dots,N_d}$ is available, where $N_d$ denotes the number of samples in the data.

Without loss of generality, in this paper we consider image data as the input, and deep convolutional neural networks as our model. In a slight abuse of notation, we also use the symbol $x$ to represent the features computed by layers within the deep network, producing a three-dimensional tensor $\mathcal X = \mathcal C \times \mathcal H \times \mathcal W$ where the dimensions indicate the number of feature channels, height and width respectively. Batch normalization (BN) computes estimators for the mini-batch $\{x_n\}_{n=1,\dots,N}$ (usually $N \ll N_d$) by average pooling over all but the channel dimensions.\footnote{How estimators are computed is what differentiates many of the normalization techniques currently available. \citet{wu18} provide a detailed introduction.} Then BN normalizes the samples in the batch as

\begin{equation}\label{eq:batch_norm}
\text{BN} (x_n) = \alpha \Big( \frac{x_n - \mu}{\sigma} \Big) + \beta,
\end{equation}

where $\mu$ and $\sigma$ are the mean and standard deviation of the mini-batch, respectively. The parameters $\alpha$ and $\beta$ are $\vert\mathcal C\vert$-dimensional vectors representing a learned affine transformation along the channel dimensions, purposed to retain each layer's representative capacity \citep{ioffe15}. This normalizing transformation ensures that the mini-batch has zero mean and unit variance when viewed along the channel dimensions.

Group normalization (GN) performs a similar transformation to that in (\ref{eq:batch_norm}), but normalizes along different dimensions. As such, GN first separates channels $c=1,\dots, \vert \mathcal C\vert$ into fixed groups $G_j$, over which it then jointly computes estimators, \eg for the mean $\mu_j = \vert G_j \vert^{-1} \sum_{x_c\in G_j} x_c$. Note that GN does not average the statistics along the mini-batch dimension, and is thus of particular interest when large batch sizes become a prohibitive factor.

A potential problem when using GN is that channels that are being grouped together might get prevented from developing distinct characteristics in feature space. In addition, computing estimators from manually engineered rules as those found in BN and GN can be too restrictive under a number of circumstances, for example when jointly learning on multiple domains.

\begin{figure}[t]
\input{mn_train}
\end{figure}
\begin{figure}[t]
\input{mn_test}
\end{figure}
\begin{figure}[t]
\input{mgn}
\end{figure}

\subsection{Mode normalization}\label{sec:methodmn}

The heterogeneous nature of complex datasets motivates us to propose a more flexible treatment of normalization. Before the actual normalization is carried out, the data is first organized into modes to which it likely belongs. To achieve this, we reformulate the normalization in the framework of mixtures of experts (MoE). In particular, we introduce a set of simple gating functions $\{g_k\}_{k=1,\dots,K}$ where $g_k \colon \mathcal X \to [0,1]$ and $\sum_k g_k(x) = 1$. In mode normalization (\mn, Alg.\ \ref{alg:mn_train}), each sample in the mini-batch is then normalized under voting from its gate assignment:

\begin{equation}\label{eq:mode_norm}
\text{MN}(x_n) \triangleq \alpha \Big( \sum_{k=1}^K g_k (x_n) \frac{x_n - \mu_k}{\sigma_k} \Big) + \beta,
\end{equation}

where $\alpha$ and $\beta$ are a learned affine transformation, just as in standard BN\@.\footnote{We experimented with learning individual $\{(\alpha_k,\beta_k)\}_{k=1,\dots,K}$ for each mode. However, we have not observed any additional gains in performance from this.}

The estimators for mean $\mu_k$ and variance $\sigma_k$ are computed under weighing from the gating network, \eg the $k$'th mean is estimated from the batch as

\begin{equation}\label{eq:mode_norm_mu}
\mu_{k} = \langle x \rangle_k = \frac{1}{N_k} \sum_{n} g_k (x_n) \cdot x_n,
\end{equation}

where $N_k = \sum_n g_k (x_n)$. In our experiments, we parametrize the gating networks via an affine transformation $\Psi \colon \mathcal X \to \mathbb R^K$ which is jointly learned alongside the other parameters of the network. This transformation is followed by a softmax activation $\sigma \colon \mathbb R^K \to [0,1]^K$, reminiscent of attention mechanisms \citep{denil12,vinyals15}. Note that if we set $K=1$, or when the gates collapse $g_k(x_n) = \text{const.}$ $\forall \,k, n$, then (\ref{eq:mode_norm}) becomes equivalent to BN, \cf (\ref{eq:batch_norm}).

As in BN, during training we normalize samples with estimators computed from the current batch. To normalize the data during inference (Alg.\ \ref{alg:mn_test}), we keep track of component-wise running estimates, borrowing from online EM approaches \citep{cappe09,liang09}. Running estimates are updated in each iteration with a memory parameter $\lambda \in (0,1]$, \eg for the mean:

\begin{equation}
\overline{\langle x \rangle}_k = \lambda \langle x \rangle_k + (1 - \lambda) \overline{\langle x \rangle}_k.
\end{equation} 

\citet{bengio15} and \citet{shazeer17} propose the use of additional losses that either prevent all samples to focus on a single gate, encourage sparsity in the gate activations, or enforce variance over gate assignments. In \mn, such additional penalties are not needed. Importantly, we want \mn to be able to seek out a form in which it recovers traditional BN, whenever that is the optimal thing to do. In practice, we seldom observed this behavior:\ gates tend to receive an even share of samples overall, and they are usually assigned to individual modes.

\subsection{Mode group normalization}

As discussed in Section~\ref{sec:background}, GN is less sensitive to the batch size \citep{wu18}. Here, we show that similarly to BN, GN can also benefit from soft assignments into different modes. In contrast to BN, GN computes averages over individual samples instead of the entire mini-batch. This makes slight modifications necessary, resulting in mode group normalization (\mgn, Alg.\ \ref{alg:mgn}). Instead of learning mappings with their preimage in $\mathcal X$, in \mgn we learn a gating network $g\colon \mathbb R \to \mathbb R^K$ that assigns channels to modes. After average-pooling over width and height, estimators are computed by averaging over channel values $x_c \in \mathbb R$, for example for the mean $\mu_k = \langle x \rangle_k = C_k^{-1} \sum_c g_k(x_c) \cdot x_c$, where $C_k = \sum_c g_k(x_c)$. Each sample is subsequently transformed via

\begin{equation}
\text{MGN}(x) \triangleq \frac{\alpha}{K} \sum_k  \frac{x - \mu_k}{\sigma_k} + \beta,
\end{equation}

where $\alpha$ and $\beta$ are learnable parameters for channel-wise affine transformations. One of the notable advantages of \mgn (that it shares with GN) is that inputs are transformed in the same way during training and inference. 

A potential risk for clustering approaches is that clusters or modes might collapse into one, as described by \eg \citet{xu2005max}. Although it is possible to address this with a regularizer, it has not been an issue in either \mn or \mgn experiments. This is likely a consequence of the large dimensionality of feature spaces that we study in this paper, as well as sufficient levels of variation in the data.

\section{Experiments}
\label{sec:experiments}

We consider two experimental settings to evaluate our methods: (i.)~multi-task, and (ii.)~single task. All experiments use standard routines within PyTorch \citep{paszke17}.

\subsection{Multi-task}

\paragraph{Data.} In the first experiment, we wish to enforce heterogeneity in the data distribution, \ie explicitly design a distribution of the form $\mathbb P = \sum_d \pi_d \mathbb P_d$. We realize this by generating a dataset whose images come from significantly diverse distributions, combining four image datasets: (i.)~\textbf{MNIST} \citep{lecun98a} which contains grayscale scans of handwritten digits. The dataset has a total of 60000 training samples, as well as 10000 samples set aside for validation. (ii.)~\textbf{CIFAR-10} \citep{krizhevsky09} is a dataset of colored images that show real world objects of one of ten classes. It contains 50000 training and 10000 test images. (iii.)~\textbf{SVHN} \citep{netzer11} is a real-world dataset consisting of 73257 training samples, and 26032 samples for testing. Each image shows one of ten digits in natural scenes. (iv.)~\textbf{Fashion-MNIST} \citep{xiao17} consists of the same number of single-channel images as are contained in MNIST\@. The images contain fashion items such as sneakers, sandals, or dresses instead of digits as object classes. We assume that labels are mutually exclusive, and train a single network\,---\,LeNet \citep{lecun89} with a 40-way classifier at the end\,---\,to jointly learn predictions on them. 

\paragraph{Mode normalization.} Training is carried out for 3.5 million data touches (15 epochs), with learning rate reductions by 1/10 after 2.5 and 3 million data touches, respectively. Note that training for additional epochs did not result in any notable performance gains. The batch size is set to $N=128$, and running estimates are kept with $\lambda=0.1$. We vary the number of modes in \mn over $K=\{2,4,6\}$. Average performances over five random initializations as well as standard deviations are shown in Table~\ref{table:mnist_and_cifar10_and_svhn_and_fashionmnist}. \mn outperforms standard BN, as well as all other normalization methods. This shows that accounting for multiple modes is an effective way to normalize intermediate features when the data is heterogeneous.

Interestingly, increasing $K$ does not always improve the performance. The reduction in effectiveness of higher mode numbers is likely a consequence of finite estimation, \ie of computing estimates from smaller and smaller partitions of the batch, a known issue in traditional BN.\footnote{Experiments with larger batch sizes support this argument, see Appendix.} In all remaining trials which involve single datasets and deeper networks, we therefore fixed $K=2$. Note that the additional overhead from coupling LeNet with \mn is limited. Even in our naive implementation, setting $K=6$ results in roughly a 5\% increase in runtime.

\begin{table}[h!]
\centering
\caption{Test set error rates (\%) of batch norm (BN), instance norm (IN) \citep{ulyanov17}, layer norm (LN) \citep{ba16}, and mode norm (MN) in the multi-task setting for a batch size of $N=128$. Shown are average top performances over five initializations alongside standard deviations. Additional results for $N=\{256,512\}$ are shown in the Appendix.}\label{table:mnist_and_cifar10_and_svhn_and_fashionmnist}
\vspace*{.4cm}
\begin{tabular}{cccccc}
\bf{BN}       & \bf{IN}       & \bf{LN}       & \bf{\mn}           & $K$   \\
\midrule
26.91 \E1.08  & 28.87 \E2.28  & 27.31 \E0.71  & \ul{23.16} \E1.23  & 2     \\
              &               &               & 24.25      \E0.71  & 4     \\
              &               &               & 25.12      \E1.48  & 6     \\
\end{tabular}
\end{table}

\paragraph{Mode group normalization.} Group normalization is designed specifically for applications in which large batch sizes become prohibitive. We therefore simulate this by reducing batch sizes to $N=\{4,8,16\}$, and train each model for \num{50000} gradient updates. This uses the same configuration as previously, except for a smaller initial learning rate $\gamma=0.02$, which is reduced by 1/10 after \num{35000} and \num{42500} updates. In GN, we allocate two groups per layer, and accordingly set $K=2$ in \mgn. As a baseline, results for BN and \mn were also included. Average performances over five initializations and their standard deviations are shown in Table~\ref{table:mnist_and_cifar10_and_svhn_and_fashionmnist_gn_mgn}. As previously shown by \cite{wu18}, BN fails to maintain its performance when the batch size is small during training. Though \mn performs slightly better than BN, its performance also degrades in this regime. GN is more robust to small batch sizes, yet \mgn further improves over GN, and\,---\,by combining the advantages of GN and \mn\,---\, achieves the best performance for different batch sizes among all four methods.

\begin{table}[h]
\centering
\caption{Test set error rates (\%) for BN, MN, mode group norm (\mgn) and group norm (GN) on small batch sizes. Shown are average top performances over five initializations alongside standard deviations.}\label{table:mnist_and_cifar10_and_svhn_and_fashionmnist_gn_mgn}
\vspace*{.4cm}
\begin{tabular}{ccccc}
$N$    & \bf{BN}        & \bf{\mn}           & \bf{GN}        & \bf{\mgn}          \\
\midrule
4      & 33.40 \E0.75   & 32.80 \E1.59       & 32.15 \E1.10   & 31.30      \E1.65  \\
8      & 31.98 \E1.53   & 29.05 \E1.51       & 28.60 \E1.45   & 26.83      \E1.34  \\
16     & 30.38 \E0.60   & 28.70 \E0.68       & 27.63 \E0.45   & \ul{26.00} \E1.68  \\
\end{tabular}
\end{table}

\subsection{Single task}

\paragraph{Data.} Here our method is evaluated in single image classification tasks, showing that it can be used to improve performance in several recently proposed convolutional networks. For this, we incorporate \mn into multiple modern architectures, first evaluating it on \textbf{CIFAR10} and \textbf{CIFAR100} datasets and later on a large-scale dataset, \textbf{ILSVRC12}~\citep{deng09}. Differently from CIFAR10, CIFAR100 has 100 classes, but contains the same number of training images, 600 images per class. ILSVRC12 contains around 1.2 million images from 1000 object categories.

\paragraph{Network In Network.} Since the original Network In Network (NIN) \citep{lin13} does not contain any normalization layers, we modify the network architecture to add them, coupling each convolutional layer with a normalization layer (either BN or \mn). We then train the resulting model on CIFAR10 and CIFAR100 for 100 epochs with SGD and momentum as optimizer, using a batch size of $N=128$. Initial learning rates are set to $\gamma=10^{-1}$, which we reduce by 1/10 at epochs 65 and 80 for all methods. Running averages are stored with $\lambda=0.1$. During training we randomly flip images horizontally, and crop each image after padding it with four pixels on each side. Dropout \citep{srivastava14} is known to occasionally cause issues in combination with BN \citep{li18}, and reducing it to 0.25 (as opposed to 0.5 in the original publication) was beneficial to performance. Note that incorporating \mn with $K=2$ into NIN adds less than 1\% to the number of trainable parameters.

We report the test error rates with NIN on CIFAR10 and CIFAR100 in Table \ref{table:cifar10_nin_resnet} (left). We first observe that NIN with BN obtains an error rate similar to that reported for the original network in~\citet{lin13}. \mn ($K=2$) achieves an additional boost of 0.4\% and 0.6\% over BN on CIFAR10 and CIFAR100, respectively.

\begin{table}[h!]
    \centering
    \caption{ Test set error rates (\%) with BN and MN for deep architectures on CIFAR10, CIFAR100. Shown are NIN (left), VGG13 (middle) and ResNet20 (right).} \label{table:cifar10_nin_resnet}
    \vspace*{.4cm}
    \begin{adjustbox}{center}
    \begin{tabular}{lcccc}
                         & \multicolumn{3}{c}{\bf{Network In Network}}           \\
    \midrule
                         & \bf{Lin et al.}  & \bf{BN}         & \bf{\mn}         \\
    \midrule
    CIFAR10              & 8.81             & 8.82            & \ul{8.42}        \\
    CIFAR100             & --               & 32.30           & \ul{31.66}       \\
    \end{tabular}
    \begin{tabular}{cc}
    \multicolumn{2}{c}{\bf{VGG13}}          \\
    \midrule
    \bf{BN}                & \bf{\mn}     \\
    \midrule
    8.28                   & \ul{7.79}    \\
    31.15                  & \ul{30.06}   \\
    \end{tabular}
    \begin{tabular}{ccc}
                           \multicolumn{3}{c}{\bf{ResNet20}}                     \\
    \midrule
                           \bf{He et al.}   & \bf{BN}         & \bf{\mn}         \\
    \midrule
                           8.75             & 8.44            & \ul{7.99}        \\
                            --              & 31.56           & \ul{30.53}       \\
    \end{tabular}
    \end{adjustbox}
\end{table}

\paragraph{VGG Networks.} Another popular class of deep convolutional neural networks are VGG networks~\citep{simonyan14}. In particular we trained a VGG13 with BN and \mn on CIFAR10 and CIFAR100. For both datasets we optimized using SGD with momentum for 100 epochs, setting the initial learning rate to $\gamma=0.1$, and reducing it at epochs 65, 80, and 90 by a factor of 1/10. The batch size is set to $N=128$. As before, we set the number of modes in \mn to $K=2$, and keep estimators with $\lambda=0.1$. When incorporated into the network, \mn improves the performance of VGG13 by 0.4\% on CIFAR10, and gains over 1\% on CIFAR100.

\paragraph{Residual Networks.} Contrary to NIN and VGG, Residual Networks \citep{he16} were originally conceptualized with layer-wise batch normalizations. We trained a ResNet20 on CIFAR10 and CIFAR100 in its original architecture (\ie with BN), as well as with \mn ($K=2$), see Table~\ref{table:cifar10_nin_resnet} (right). On both datasets we follow the standard training procedure and train both models for 160 epochs, with SGD as optimizer, momentum parameter of 0.9, and weight decay of \num{e-4}. Running estimates were kept with $\lambda=0.1$, the batch size set to $N=128$. Our implementation of ResNet20 (BN in Table~\ref{table:cifar10_nin_resnet}) performs slightly better than that reported in the original publication (8.42\% versus 8.82\%). Replacing BN with \mn achieves a notable 0.45\% and 0.7\% performance gain over BN in CIFAR10 and CIFAR100, respectively.

We also test our method in the large-scale image recognition task of ILSVRC12. Concretely, we replaced BN in a ResNet18 with \mn ($K=2$), and trained both resulting models on ILSVRC12 for 90 epochs. We set the initial learning rate to $\gamma=0.1$, reducing it at epochs 30 and 60 by a factor of 1/10. SGD was used as optimizer (with momentum parameter set to 0.9, weight decay of \num{e-4}). To accelerate training we distributed the model over four GPUs, with an overall batch size of $N=256$. As can be seen from Table~\ref{table:imagenet_resnet}, \mn results in a small but consistent improvement over BN in terms of top-1 and top-5 errors.

\begin{table}[h!]
    \centering
    \caption{Top-1 and top-5 error rates (\%) of ResNet18 on ImageNet ILSVRC12, with BN and \mn.} \label{table:imagenet_resnet}
    \vspace*{.4cm}
    \begin{tabular}{ccc}
        \bfseries{Top-$k$ Error} & \bfseries{BN}      & \bfseries{\mn} \\
        \midrule
        1                        & 30.25              & \ul{30.07}     \\
        5                        & 10.90              & \ul{10.65}     \\
    \end{tabular}
\end{table}

\begin{figure*}[h]
\begin{center}
\texttt{conv3-64-1} \\
\vspace{.15cm}
\includegraphics[width=.38\textwidth]{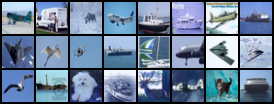}
\hspace{.02cm}
\includegraphics[width=.38\textwidth]{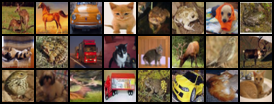}
\\
\texttt{conv3-256-1} \\
\vspace{.15cm}
\includegraphics[width=.38\textwidth]{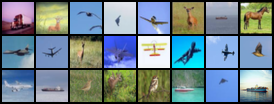}
\hspace{.02cm}
\includegraphics[width=.38\textwidth]{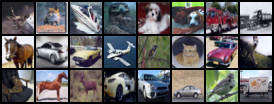}
\end{center}
\caption{Test samples from CIFAR10 that were clustered together by two experts in an early layer (top) and a deeper layer (bottom) of VGG13.} \label{fig:vgg_samples}
\end{figure*}

\paragraph{Qualitative analysis. } In Fig.~\ref{fig:vgg_samples} we evaluated the experts $g_k(\{x_n\})$ for samples from the CIFAR10 test set in layers \texttt{conv3-64-1} and \texttt{conv-3-256-1} of VGG13, and show those samples that have been assigned the highest probability to belong to either of the $K=2$ modes. In the normalization belonging to \texttt{conv3-64-1}, \mn is sensitive to a red-blue color mode, and separates images accordingly. In deeper layers such as \texttt{conv-3-256-1}, separations seem to occur on the semantic level. In this particular example, \mn separates smaller objects from such that occupy a large portion of the image.

\section{Conclusion}
\label{sec:conclusion}

Stabilizing the training process of deep neural networks is a challenging problem. Several normalization approaches that aim to tackle this issue have recently emerged, enabling training with higher learning rates, faster model convergence, and allowing for more complex network architectures.

Here, we showed that two widely used normalization techniques, BN and GN, can be extended to allow the network to jointly normalize its features within multiple modes. We further demonstrated that our method can be incorporated to various deep network architectures and improve their classification performance consistently with a negligible increase in computational overhead. As part of future work, we plan to explore customized, layer-wise mode numbers in \mn, and automatically determining them, \eg by utilizing concepts from sparsity regularization.

\bibliography{literature}
\bibliographystyle{iclr2019_conference}

\begin{appendix}
\label{sec:appendix}

\section{Additional multi-task results}

Shown in Table~\ref{table:mnist_and_cifar10_and_svhn_and_fashionmnist_appendix} are additional results for jointly training on MNIST, CIFAR10, SVHN, and Fashion-MNIST. The same network is used as in previous multi-task experiments, for hyperparameters see Section \ref{sec:experiments}. In these additional experiments, we varied the batch size to $N=\{256, 512\}$. For larger batch sizes, increasing $K$ to values larger than two increases performance, while for a smaller batch size of $N=128$ (\cf Table~\ref{table:mnist_and_cifar10_and_svhn_and_fashionmnist}), errors incurred by finite estimation prevent this benefit from appearing.

\begin{table}[h]
\centering
\caption{Test set error rates (\%) of multiple normalization methods in the multi-task setting for large batch sizes. The table contains average performances over five initializations, alongside their standard deviation.}\label{table:mnist_and_cifar10_and_svhn_and_fashionmnist_appendix}
\vspace*{.4cm}
\begin{tabular}{ccccccc}
    $N$ & \bf{BN}      & \bf{IN}        & \bf{LN}       & \bf{\mn}         & $K$\\
    \midrule
    256 & 26.34 \E1.82 &  31.15 \E3.45  & 26.95 \E2.51  & {25.29}   \E1.31 & 2 \\
        &              &                &               & 25.04     \E1.88 & 4 \\
        &              &                &               &\ul{24.88} \E1.24 & 6 \\
    \midrule
    512 & 26.51 \E1.15   & 29.00 \E1.85   & 28.98 \E1.32   & 26.18 \E1.86     & 2 \\
        &                &                &                &\ul{24.29} \E1.82 & 4 \\
        &                &                &                & 25.33 \E1.33     & 6 \\
\end{tabular}
\end{table}
\end{appendix}

\end{document}